\documentclass[runningheads]{llncs}
\usepackage{graphicx}
\usepackage[table,dvipsnames]{xcolor} %for use in color links 
\usepackage{color}
\usepackage[]{todonotes}

\usepackage{tikz}
\usetikzlibrary{%
				automata,%
				arrows,%
				backgrounds,%
				calc,%
				chains,%
				decorations,%
				decorations.markings,%
				decorations.pathmorphing,%
				external,%
				fadings,%
				fit,%
				matrix,%
				positioning,%
				scopes,%
				shadows,%
				shapes.geometric,%
				shapes.misc,%
				shapes.multipart,%
				through,%
				mindmap, %added 
				backgrounds, %added% drawing the background after the foreground
				topaths}
\usepackage{tikz-3dplot}
\usepackage{units}%nicefrac

\usepackage[colorlinks=true,linkcolor=black,anchorcolor=black,citecolor=black,filecolor=black,menucolor=black,runcolor=black,urlcolor=black]{hyperref}
\usepackage{amsmath, amsfonts, amssymb}
\definecolor{KITred}{RGB}{160,30,40}

\definecolor{KITblue}  {RGB}{ 70,100,170} % 100% blue
\definecolor{KITblue70}{RGB}{125,146,195} %  70% blue
\definecolor{KITblue50}{RGB}{162,177,212} %  50% blue
\definecolor{KITblue30}{RGB}{199,208,229} %  30% blue
\definecolor{KITblue15}{RGB}{227,232,242} %  15% blue
\definecolor{KITlilac}{RGB}{160,0,120}

\begin{document}
\title{Handling Missing Observations with an {RNN}-based Prediction-Update Cycle}

%\author{Anonymous submission}


%arxiv
\author{Stefan Becker \inst{1}\href{https://orcid.org/0000-0001-7367-2519}{\includegraphics[scale=0.04]{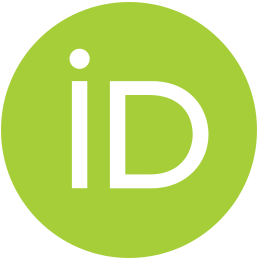}} \and
Ronny Hug \inst{1}\href{https://orcid.org/0000-0001-6104-710X}{\includegraphics[scale=0.04]{images/orcid-og-image.png}} \and 
Wolfgang Huebner \inst{1}\href{https://orcid.org/0000-0001-5634-6324}{\includegraphics[scale=0.04]{images/orcid-og-image.png}}   \and \\ Michael Arens \inst{1}\href{https://orcid.org/0000-0002-7857-0332}{\includegraphics[scale=0.04]{images/orcid-og-image.png}}
\and Brendan T. Morris\inst{2}\href{https://orcid.org/0000-0002-8592-8806}{\includegraphics[scale=0.04]{images/orcid-og-image.png}} } 

\authorrunning{S. Becker et al.}
\institute{Fraunhofer IOSB\thanks{Fraunhofer IOSB is a member of the Fraunhofer Center for Machine Learning.}, Ettlingen, Germany\\
\email{\{firstname.lastname\}@iosb.fraunhofer.de}\\
\url{www.iosb.fraunhofer.de} \and
University of Nevada, Las Vegas, USA\\
\email{brendan.morris@unlv.edu}}
\maketitle              % typeset the header of the contribution
\begin{abstract}
In tasks such as tracking, time-series data inevitably carry missing observations. While traditional tracking approaches can handle missing observations, \emph{recurrent neural networks} (RNNs) are designed to receive input data in every step. Furthermore, current solutions for RNNs, like omitting the missing data or data imputation, are not sufficient to account for the resulting increased uncertainty.
Towards this end, this paper introduces an RNN-based approach that provides a full temporal filtering cycle for motion state estimation. The Kalman filter inspired approach, enables to deal with missing observations and outliers. For providing a full temporal filtering cycle, a basic RNN is extended to take observations and the associated belief about its accuracy into account for updating the current state. An RNN prediction model, which generates a parametrized distribution to capture the predicted states, is combined with an RNN update model, which relies on the prediction model output and the current observation. By providing the model with masking information, binary-encoded missing events, the model can overcome limitations of standard techniques for dealing with missing input values. The model abilities are demonstrated on synthetic data reflecting prototypical pedestrian tracking scenarios.

\keywords{Recurrent Neural Networks (RNNs) \and Trajectory Data \and Missing Input Data \and Outliers \and Filtering.}
\end{abstract}

\section{Introduction \& Related Work}
\label{sec:intro}
One important task for autonomous systems is estimating pedestrians' motion states based on observations. After the success of RNNs in a variety of sequence processing tasks, like speech recognition \cite{chung2015recurrent,Graves_ICASSP_2013} and caption generation \cite{Donahue_CVPR_2015,Xu_MLR_2015}, these models are also successfully applied to pedestrian trajectory prediction (see for example \cite{Alahi_CVPR_2016,Hasan_CVPR_2018,Hug_RFMI_2017,Syed_AVC_2020}). While tracking approaches based on Bayesian formalization explicitly model the increase in the prediction uncertainty when an observation is missing, RNN-based models are designed to receive input data in every step. The two main ways to address missing values in time series are data imputation and omitting the missing data \cite{Schafer_PM_2002}. Data imputation means to substitute the missing values with methods like interpolation \cite{Kreindler_2006} or spline fitting \cite{DeBoor_2001}. Nonetheless, various imputation methods estimate better missing data, which results in a process where imputation and prediction models are separated \cite{Che_SREP_2018}. Since the model does not effectively explore the missing pattern, only suboptimal results are achieved. The simplest omitting strategy is to remove samples in which a value is missing. This may work for training but cannot be applied during inference. Alternatively, and in particular for RNNs, the problem can be modeled with marked missing values. A missing value can be masked and explicitly excluded, or the model can be encouraged to learn that a specific value represents the missing observation \cite{brownlee2017}. 
Most approaches are for healthcare applications \cite{Tresp_NIPS_1997} or in the field of speech recognition \cite{Parveen_NIPS_2001}. More recently, Che et al.\cite{Che_SREP_2018} customized an RNN model to incorporate the patterns of missingness for time series classification. Also, for classifying time series, Lipton et al. \cite{Lipton_PMLR_2016} treated the pattern of missing data as a feature to diagnose clinical data collected from a pediatric intensive care unit.\\
This paper introduces an RNN-based full temporal filtering cycle for motion state estimation to better deal with missing observations. The approach is intended to serve as a module for single object motion filtering in a multi-object deep learning trajectory prediction pipeline. In trajectory prediction applications, deep learning-based approaches are increasingly replacing classic approaches due to their ability to better capture contextual cues from the static (e.g. obstacles) or dynamic environment (e.g. other objects in the scene) \cite{Rudenko_IJRR_2020}. Although there exist variants relying on \emph{generative adversarial networks} (GANs) \cite{Amirian_CVPR_W_2019,Gupta_CVPR_2018}, \emph{temporal convolution networks} (TCNs) \cite{Becker_ECCVW_2018,Nikhil_ECCVW_2018}, and \emph{transformers} \cite{Giuliari_ICPR_2020,Saleh_arXiv_2020}, for encoding object motion, the most popular basis is RNNs. 
The proposed approach can partly be adapted to the other deep learning approaches but is then essentially limited to the additional masking information. It should be noted that due to the positional encoding and the attention mechanism applied in \emph{transformers}, these models can deal with missing observations by exploiting the remaining observations \cite{Giuliari_ICPR_2020}. The positional encoding extends to unseen lengths, but it is primarily designed for a fixed input length. Thus, it is not clear how well this approach generalizes to variable input lengths, and our proposed approach is designed for varying input lengths. For a comprehensive overview of current deep learning-based approaches for trajectory prediction, the reader is referred to these surveys \cite{Rasouli_arXiv_2020,Rudenko_IJRR_2020,Kothari_arXiv_2020}.\\ 
For providing a full temporal filtering cycle, two RNNs are combined to recursively infer the prior and posterior motion states. Thereby, an RNN update model (Update-RNN) relies on the output of an RNN prediction model (Prediction-RNN) in addition to the current observation for inferring the current state. The Prediction-RNN generated a parametrized distribution to capture future states and their prediction uncertainties. Both networks are additionally provided with masking information to enable the networks to learn a representation for missing observations. Thereby, the Prediction-RNN can capture the increased uncertainty when observations are missing, and the Update-RNN can learn to trust in the prior states in these situations. The evaluation is done on synthetically generated data reflecting prototypical pedestrian tracking scenarios.\\
In the following, a brief formalization of the problem and a description of the RNN-based Prediction-Update-Cycle are provided in section \ref{sec:model}. The achieved results are presented in section \ref{sec:Evaluation}.
Finally, a conclusion is given in section \ref{sec:conclusion}.
\section{RNN-based Prediction-Update-Cycle}
\vspace{-4.0mm}
\label{sec:model}
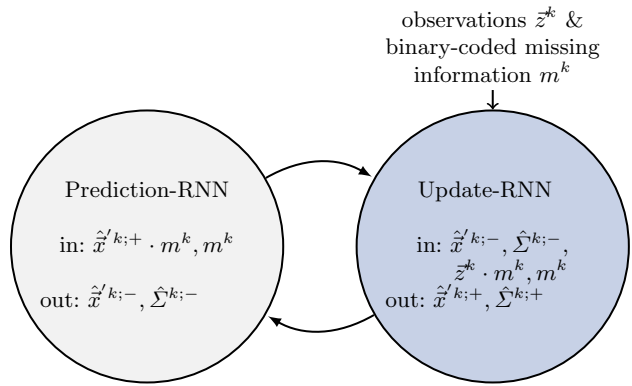
\begin{figure}[h]%
  \centering%
  \tikzsetnextfilename{red_predictor}%
  \resizebox{0.7\textwidth}{!}{%   <--- optionale Skalierung
    \resizebox{0.8\columnwidth}{!}{
\begin{tikzpicture}
\begin{scope}[font=\footnotesize]%
 % Setup the style for the states
			%\tikzset{node style/.style={state, scale=2.,minimum width={width("Prediction")+2pt}}}	
			\tikzset{state/.style={circle, draw, minimum size=3cm, line width=0.8pt}}
      % \draw[help lines, white] (-2,-2) grid (2,2);
      % Draw the states
			\node[state,fill=gray!10,text width=3.0cm]  at(-2,0) (s) {\centering Prediction-RNN \\\text{  } \\in: $\hat{\vec{x}}^{{'k;+}} \cdot {m}^{k}, {m}^{k}$ \\
			 \text{    }\\
			out: $\hat{\vec{x}}^{'k;-}, \hat{\Sigma}^{k;-}$ };
			%\node[rectangle, draw, minimum height=1cm, minimum width=1cm] (Y0) at (-1, 0) { };
			%\node[rectangle, rounded corners=10, minimum width=2cm, minimum height=1cm, draw, very thick] (lstm) at (-2, 0) {};
			%\node[rectangle, draw, right= of Y0, minimum height=1cm, minimum width=1cm] (RNN) {$\mathbf{h}^{t-1}_{1}$};			
			\node[state, right=of s,fill=KITblue!30,text width=3.0cm] (r) {\centering Update-RNN \text{  } \\ \text{  }\\
			in: $\hat{\vec{x}}^{'k;-}, \hat{\Sigma}^{k;-},  $   \\
			\hspace{0.4cm} ${\vec{z}}^{k}\cdot{m}^{k}, {m}^{k}   $  \\
			out: $\hat{\vec{x}}^{'k;+}, \hat{\Sigma}^{k;+}$};				
			\node[text width=3.5cm,align=center] (obs) [above=2ex of r]{observations $\vec{z}^{k}$ \& \\ binary-coded missing information ${m}^{k}$};			
			%\node(init)[text width=3cm,align=center] at (-3,1.5) {Initial conditions};
			%\node(init)[text width=3cm,align=center] [above=2ex of s]{Initial conditions};				
			%\draw[ultra thick,loosely dashed,white] (2,0) -- (2,4);				
		 \draw[thick,->] (obs) -- (r.north) ;
		% \draw[thick,->] (init) -- (s.north) ;				
			% Connect the states with arrows
			\draw[every loop, >=latex]
					(s) edge[bend left, thick] node {} (r)
					(r) edge[bend left, thick] node {} (s);						
				 %(s) edge[loop above] node {} (s)
				 %(r) edge[loop above] node {} (r)	;	
			%	\draw[help lines] (-3,-3) grid (3,3);
\end{scope}				
\end{tikzpicture}
}%
  }%                           <--- optionale Skalierung
	\vspace{-1.0mm}
  \caption[]{Visualization of the proposed RNN-based \emph{prediction-update} cycle. The Update-RNN estimates the unknown system state $\hat{\vec{x}}^{'k}$ from the observations $\vec{z}^k$ and estimated prior state $\hat{\vec{x}}^{'k-1}$ or rather $\hat{\vec{x}}^{'k;-}$ provided by the Prediction-RNN.}		
	\label{fig:rnn_pred_update_cycle}%
\end{figure}

The goal is to devise a model that can successfully infer motion states of tracked objects and deal with missing observations. In the context of RNNs, trajectory prediction is formally stated as the problem of inferring trajectories of objects (e.g. pedestrians), conditioned on their track history. Given an input sequence $\mathcal{Z}$ of consecutive observed positions $\vec{z}^k=(x^k,y^k)$ at time step $k$ along a trajectory, the task is to filter the current position $\hat{\vec{z}}^k=(x^k,y^k)$ and to generate predictions for future positions $\{ \vec{z}^{k+1}, \vec{z}^{k+2},\ldots \}$. Almost all deep-learning-based trajectory prediction models conditioned solely on positions ignore that the observed positional data includes uncertainties \cite{Giuliari_ICPR_2020}. Conditioning is done under the assumption that a noise-free, full input trajectory is provided. We combine two RNNs in a Kalman filter-like Prediction-Update cycle to deal with the included uncertainties in the observations.\\
 
\textbf{Prediction Network:} The Prediction-RNN generates the distribution over the next position $\vec{z}^{k+1}$, the density of the predicted state $ p^{-}((\vec{z}=\vec{x}')^{k+1}) \triangleq p^{}_{}( \vec{x}'^{k+1} | \vec{z}^{ 0:k} )$. Compared to Bayesian filtering, $\vec{x}'^{k}$ is not the full dynamical state $\vec{x}^{k}$, but the state $\vec{z}^k$ can be interpreted as observable state by mapping the RNN state $\vec{h}$ to the observation space \cite{Becker_phd_thesis_2020}. For generating the distribution $p^{-}(\cdot )$ over the next positions, the model parametrizes a mixture density network (MDN)\cite{Bishop_techreport_1994}. For reflecting the increased prediction uncertainty in case of a missing observation, which would result in a changed possible position distribution, we propose to extend the input sequence with masking information in the form of a binary-coded indicator variable ${m}^{k}\in \{0,1\}$, which marks an observation as missing. The binary-coded masking is used to incorporate a replacement value as missing. In the case of dealing with missing last $k_{miss}$ observation, the model generates a distribution over the $1+k_{miss}$ next steps. In practice, only conditioned on the information from the $k-k_{miss}$ observed time steps. The model can be trained by maximizing the likelihood of the data given the output Gaussian mixture parameters. The loss function $\mathcal{L}_{pred}$ of the Prediction-RNN using one mixture component is given by $\mathcal{L}_{pred} ( \mathcal{Z}')  =  \sum_{k=1}^K -\log \left\{ \mathcal{N}(\vec{z}^{k+1+k_{miss}} | \vec{x}^{'k}, \Sigma^{k}) \right\}$.
Thus, for $k_{miss}=0$ this is the default loss for a next step prediction RNN-MDN. $\mathcal{Z}'$ is the combination of estimated states by the Update-RNN multiplied with the masking information and concatenation of the masking information $\{\hat{\vec{x}}^{'k} \cdot {m}^{k}, {m}^{k}\}$. Although the Update-RNN estimates the current state $\vec{x}'^{k}$ for every time step, the estimates are replaced with the missing placeholder for conditioning. Note that the Prediction-RNN is not used for long-term prediction but for providing the Update-RNN with a prior state with uncertainty $\hat{\vec{x}}^{'k;-}, \hat{\Sigma}^{k;-}$ together with observations. Since RNNs are only capable of generating conditional predictions for one time step at a time, we can create a next step prediction with increased uncertainty using the missing placeholders. With an embedding of the  inputs, the Prediction-RNN can be defined as follows:
\begin{align}
\centering
 \vec{e}^{k}_{pred} = \text{EMB}&(\hat{\vec{x}}^{{'k;+}} \cdot {m}^{k}, {m}^{k}; \vec{\Theta}_{epred} ) \text{,} \nonumber \\
 \vec{h}^{k}_{pred} = \text{RNN}&(\vec{h}^{k-1}_{pred},\vec{e}^{k}_{pred}; \vec{\Theta}_{RNNpred} ) \text{,}\nonumber \\ 
 \hat{\vec{z}}^{k+1+k_{miss}}, \hat{{\Sigma}}^{k+1+k_{miss}} &= \text{MLP}(\vec{h}^{k}_{pred}; \vec{\Theta}_{MLPpred}) \nonumber \\ 
\hat{\vec{x}}^{'k;-}, \hat{\Sigma}^{k;-} &= \hat{\vec{z}}^{k+1+k_{miss}}, \hat{{\Sigma}}^{k+1+k_{miss}} 
\label{eq:RNN_pred} 
\end{align}

Here, $\text{RNN}(\cdot)$ is the recurrent network, $\vec{h}$ the hidden state of the RNN, $\text{MLP}(\cdot)$ the multilayer perceptron, and $\text{EMB}(\cdot)$ an embedding layer. $\vec{\Theta}$ represents the parameters (weights and biases) of the MLP, EMB or respectively RNN.\\

\textbf{Update Network:} The Update-RNN is used for generating the posterior $ p^{+}(\vec{x}^{'k})$. The posterior is the probability distribution over $\vec{x}^{'k}$ conditioned on all past observations $\vec{z}^{0:k}$. It is important to note, that the $\hat{\vec{x}}^{'k;+}, \hat{\Sigma}^{k;+}$ and $\hat{\vec{x}}^{'k;-}, \hat{\Sigma}^{k;-}$ depend on the whole history of inputs in contrast to the Markov assumption of the Kalman filter. In case of a missing observation, the corresponding $\vec{z}^{k}$ is replaced with a placeholder value by multiplying with the masking value $m^{k}$. Here, we used zero as placeholder values for missing. Besides the observations, the output of the Prediction-RNN is also used as input for the Update-RNN. Although a division is not defined for matrices (Kalman gain multiplies prior uncertainty with the inverse observation uncertainty), we can think of the Kalman gain as a ratio that controls the influence of a new observation on the updated (posterior) state estimate. Following a Kalman filter, the Update-RNN learns to weight both inputs in order to generate the parameter of an MDN for representing the posterior. The weighting factors $K$ ($K_{pred}=(1-K_{obs})$) can therefore be seen as a pseudo-Kalman gain. The Update-RNN with an embedding of the inputs is given by:
\begin{align}
\centering
 \vec{e}^{k}_{up} = \text{EMB}(\hat{\vec{x}}^{'k;-}, \hat{\Sigma}^{k;-}, &\tilde{\vec{z}}^{k} \cdot {m}^{k}  , {m}^{k} ; \vec{\Theta}_{eup} ) \text{,} \nonumber \\
 \vec{h}^{k}_{up} = \text{RNN}(\vec{h}^{k-1}_{up},\vec{e}^{k}_{up}; &\vec{\Theta}_{RNNup} ) \text{,}\nonumber \\ 
 \hat{\vec{z}}^{k}=K_{pred} \cdot\hat{\vec{x}}^{'k;-}+ K_{obs} \cdot \tilde{\vec{z}}^{k}, \hat{{\Sigma}}^{k} &= \text{MLP}(\vec{h}^{k}_{up}; \vec{\Theta}_{MLPup}) \nonumber \\ 
\hat{\vec{x}}^{'k;+}, \hat{\Sigma}^{k;+} = \hat{\vec{z}}^{k}, &\hat{{\Sigma}}^{k} 
\label{eq:RNN_up} 
\end{align}
	
Here, $\tilde{\vec{z}}^{k}$ is an actual, noisy observation, a realization of $\vec{z}^k$ despite the inputs of an RNN being deterministic. The pseudo-Kalman gain can be realized with every activation function keeping the output between zero and one, switching between trusting the prior or the current observations. Here, K is generated with a \emph{softplus} activation function. The Update-RNN can learn when to rely on predictions instead of observations due to provided prediction uncertainty and the masking information. Similar to Kalman filtering, the information of both RNN is exchanged iteratively. The Update-RNN is trained by minimizing the filtering loss $\mathcal{L}_{up}$ in the form of the negative log-likelihood of the ground truth current position under the filtered position. By combining both models, we get a full Prediction-Update cycle to filter noisy observations and handle missing observations from variable input sequences. In figure \ref{fig:rnn_pred_update_cycle} the RNN-based Prediction-Update cycle is visualized.            

\section{Data Generation and Evaluation}
\label{sec:Evaluation}
This section consists of a brief evaluation of the proposed Prediction-Update-RNN cycle. The evaluation is concerned with verifying the approach's overall viability in situations with missing observations and outliers from tracking maneuvering pedestrians. For initial results, synthetic generated data is used due to the fact that current pedestrian trajectory data sets do not consider aspects like motion smoothness (see for example \emph{TrajNet++} \cite{Kothari_arXiv_2020}, \emph{UCY} \cite{Lerner_CGF_2007},\emph{ETH} \cite{Pellegrini_ICCV_2009},  SDD \cite{Robicquet_ECCV_2016}) despite RNNs can generalize to deal with noisy inputs. Further, problems such as limited training samples are avoided. Although using synthetic data, we make use of a real-world dataset with maneuvering pedestrians to capture similar conditions (\emph{Daimler Path Prediction} dataset \cite{Schneider_GCPR_2013}). For generating synthetic trajectories of a basic maneuvering pedestrian on a ground plane, random agents are sampled from a Gaussian distribution according to a preferred pedestrian walking speed \cite{Teknomo_phd_thesis_2002} ($\mathcal{N}(1,38\nicefrac{m}{s}, (0.37\nicefrac{m}{s})^2 )$). The frame rate is set to $16{fps}$. During a single trajectory simulation, the agents can perform a turning maneuver. The heading change is sampled from a uniform distribution between $45^{\circ}$ and $100^{\circ}$. The duration of the turning event is sampled from a Gaussian distribution based on the mean sojourn time estimated from the ground truth sequences ($\mathcal{N}(1.83{s}, (0.29{s})^2)$). The positional observation noise is assumed to follow a bimodal Gaussian mixture model for considering outliers. The outlier observation noise is set to $\sigma_{outl}=0.5{m}$ and the standard observation noise is varied ($\sigma_{w}=0.05{m}$ and $\sigma_{w}=0.01{m}$). Outlier and missing events are drawn from a Bernoulli distribution $Ber(\cdot,\cdot)$.\\ 
\begin{table*}[t!]
\caption{Results for a comparison between the proposed RNN-based Prediction-Update cycle compared to two variants of RNN-MDNs ($1$to$1$ and encoder). The displacement error is shown for different observation noise levels and for varying probabilities of outliers and missing observations.}
	 % \begin{center}
	%%\resizebox{\columnwidth}{!}{%
	\resizebox{\textwidth}{!}{
    \begin{tabular}{| c | c c | c c | c c | c c |}
      \hline
      \hline
			\multicolumn{1}{|c|}{} & \multicolumn{8}{c|}{ fully observed $Ber_{miss}(0.0,1.0)$} \\
			\hline
			\multicolumn{1}{|c|}{} & \multicolumn{4}{c|}{no outlier $Ber_{outl}(0.0,1.0)$} & \multicolumn{4}{c|}{with outlier ($Ber_{outl}(0.1,0.9)$;$\sigma_{outl}=0.5$)}  \\	
			 \hline
      %\makebox[10mm]{Table} & \makebox[10mm]{1st} &   \makebox[10mm]{2nd} & \makebox[10mm]{3rd} & \makebox[10mm]{4th} &\makebox[10mm]{5th} &   \makebox[10mm]{6th} & \makebox[10mm]{7th} & \makebox[10mm]{8th}\\
			\multicolumn{1}{|c|}{Approach} & \multicolumn{2}{c|}{ $\sigma_{w}=0.01$} &\multicolumn{2}{c|}{$\sigma_{w}=0.05$}  & \multicolumn{2}{c|}{$\sigma_{w}=0.01$} & \multicolumn{2}{c|}{$\sigma_{w}=0.05$}    \\
      %\hline
			\multicolumn{1}{|c|}{} & {ADE/m}&$\sigma_{ADE}$/m & {ADE/m}&$\sigma_{ADE}$/m  & {ADE/m}&$\sigma_{ADE}$/m &{ADE/m}&$\sigma_{ADE}$/m    \\
			\hline
			%\multicolumn{1}{|c|}{} & \multicolumn{1}{c}{ADE / m} & $\sigma_{\text{ADE}}$ / m & \multicolumn{1}{c}{ADE / m} & $\sigma_{\text{ADE}}$ / m  & ADE / m    &   \multicolumn{1}{c|}{$\sigma_{\text{ADE}}$ / m } & \multicolumn{1}{c}{ADE / m} & $\sigma_{\text{ADE}}$ / m  \\
			\rowcolor{red!15}
      Prediction-Update-RNN & 0.011 &  0.012   & 0.051 & 0.027   & 0.038 & 0.097   & 0.083 &0.106  \\
			\rowcolor{KITblue!30}
      RNN-(1to1)-MDN & 0.018   & 0.039 & 0.053 & 0.064  & 0.076 & 0.084   & 0.094   & 0.096\\
			\rowcolor{KITblue!30}
      RNN-(encoder)-MDN  & 0.028   & 0.016   & 0.067 &  0.035  & 0.112 & 0.214   & 0.135 &  0.196\\
      \hline
			\hline
			\multicolumn{1}{c}{} & \multicolumn{8}{c}{} \\
			  \hline
			\hline
			\multicolumn{1}{|c|}{} & \multicolumn{8}{c|}{missing observations $Ber_{miss}(0.1,0.9)$} \\
			\hline
			\multicolumn{1}{|c|}{} & \multicolumn{4}{c|}{no outlier $Ber_{outl}(0.0,1.0)$} & \multicolumn{4}{c|}{with outlier ($Ber_{outl}(0.1,0.9)$;$\sigma_{outl}=0.5$) }  \\	
			 \hline
			\multicolumn{1}{|c|}{Approach} & \multicolumn{2}{c|}{ $\sigma_{w}=0.01$} &\multicolumn{2}{c|}{$\sigma_{w}=0.05$}  & \multicolumn{2}{c|}{$\sigma_{w}=0.01$} & \multicolumn{2}{c|}{$\sigma_{w}=0.05$}    \\
      %\hline
			\multicolumn{1}{|c|}{} & {ADE/m}&$\sigma_{ADE}$/m & {ADE/m}&$\sigma_{ADE}$/m  & {ADE/m}&$\sigma_{ADE}$/m &{ADE/m}&$\sigma_{ADE}$/m    \\
			\hline
			\rowcolor{red!15}
			 Prediction-Update-RNN  &  0.021 & 0.057    & 0.060 & 0.058  & 0.039 & 0.096   & 0.090  & 0.114   \\
			\rowcolor{KITblue!30}
			 RNN-(1to1)-MDN (imputation) & 0.031   & 0.056   &  0.065   & 0.080    & 0.087  & 0.110   & 0.101  & 0.120  \\
			\rowcolor{KITblue!30}
			 RNN-(encoder)-MDN (imputation) & 0.040 &  0.029   & 0.069 & 0.037  &  0.104 &  0.194    & 0.138 &  0.195 \\			
      \hline
			\hline
    \end{tabular}
    \label{result-table-pred-update-cycle}
  %\end{center}
	}
\end{table*}

\textbf{Implementation Details:} The models have been implemented using \emph{Pytorch} \cite{Paszke_NEURIPS_2019}. The Prediction-RNN is pre-trained for $100$ epochs on noise-free trajectory data and then for $100$ epochs on noisy trajectory data. After that, both models are jointly trained for $400$ epochs. In the joint training, the estimated states are iteratively exchanged over the sequence length, whereas in pre-training, the Prediction-RNN is conditioned directly on the observations. For training, the ADAM optimizer \cite{Kingma_ICLR_2015} with a learning rate of $0.001$ is used. As RNN variant, the \emph{long short-term memory} (LSTM) \cite{Hochreiter_NC_1997} is utilized. 
%\vspace{-5.0mm}
\begin{figure}[!ht]
\centering
\resizebox{\textwidth}{!}{
\begin{tabular}{cc}				
				\begin{tikzpicture}
					\begin{scope}[font=\footnotesize]
					\node (image1) at (0,0) {	\includegraphics[width=70mm]{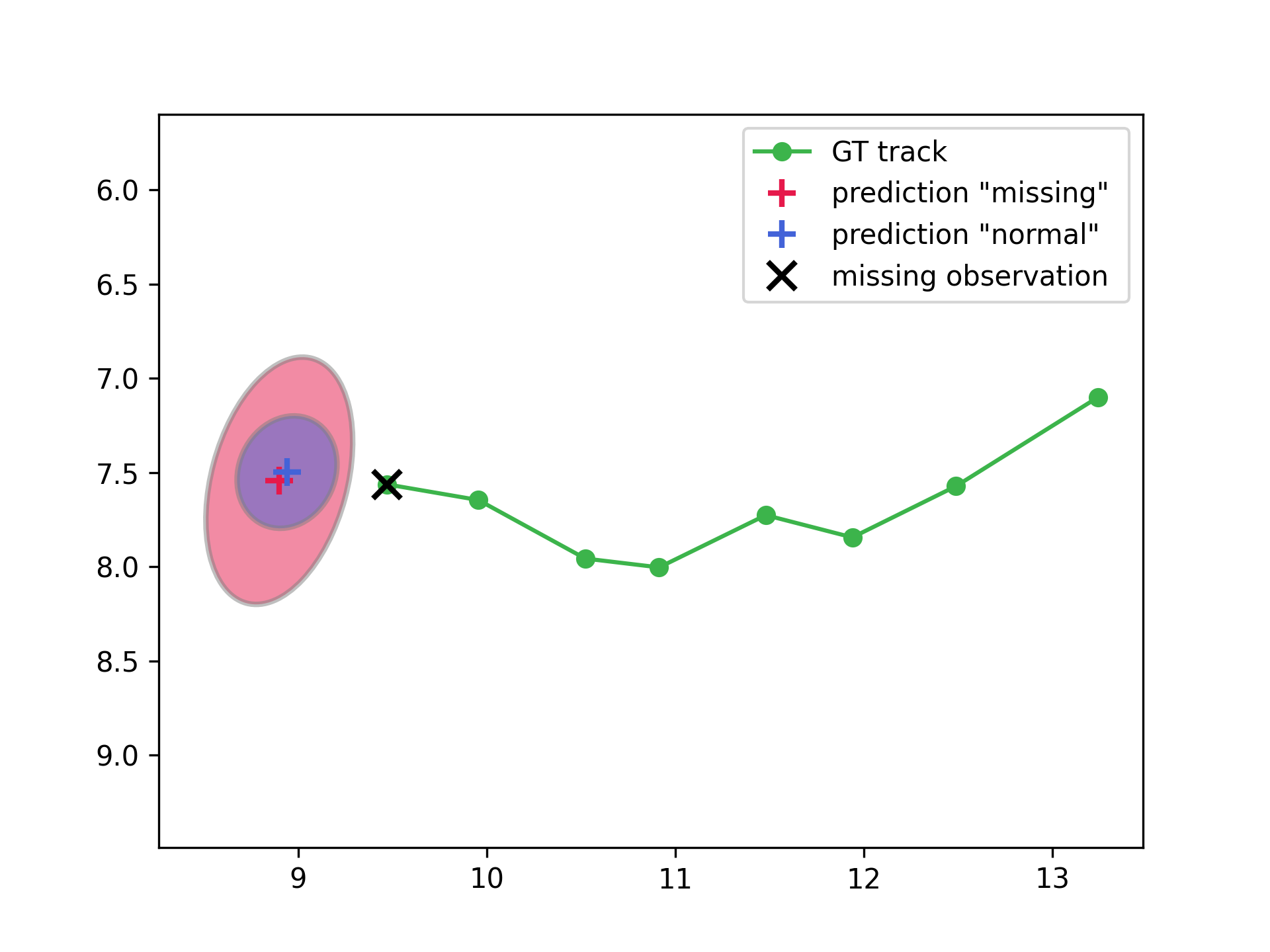}};			
					\node [below of = image1,node distance=2.5cm] (x){$x$/m};				
					\node [left of = image1, node distance=3.3cm,rotate=90] (Y) {$y$/m};
					%\draw[step=0.5cm,gray,very thin] (-4,-4) grid (4.0,4.0);
					\end{scope}									
				\end{tikzpicture} &
				\begin{tikzpicture}
					\begin{scope}[font=\footnotesize]
					\node (image1) at (0,0) {	\includegraphics[width=70mm]{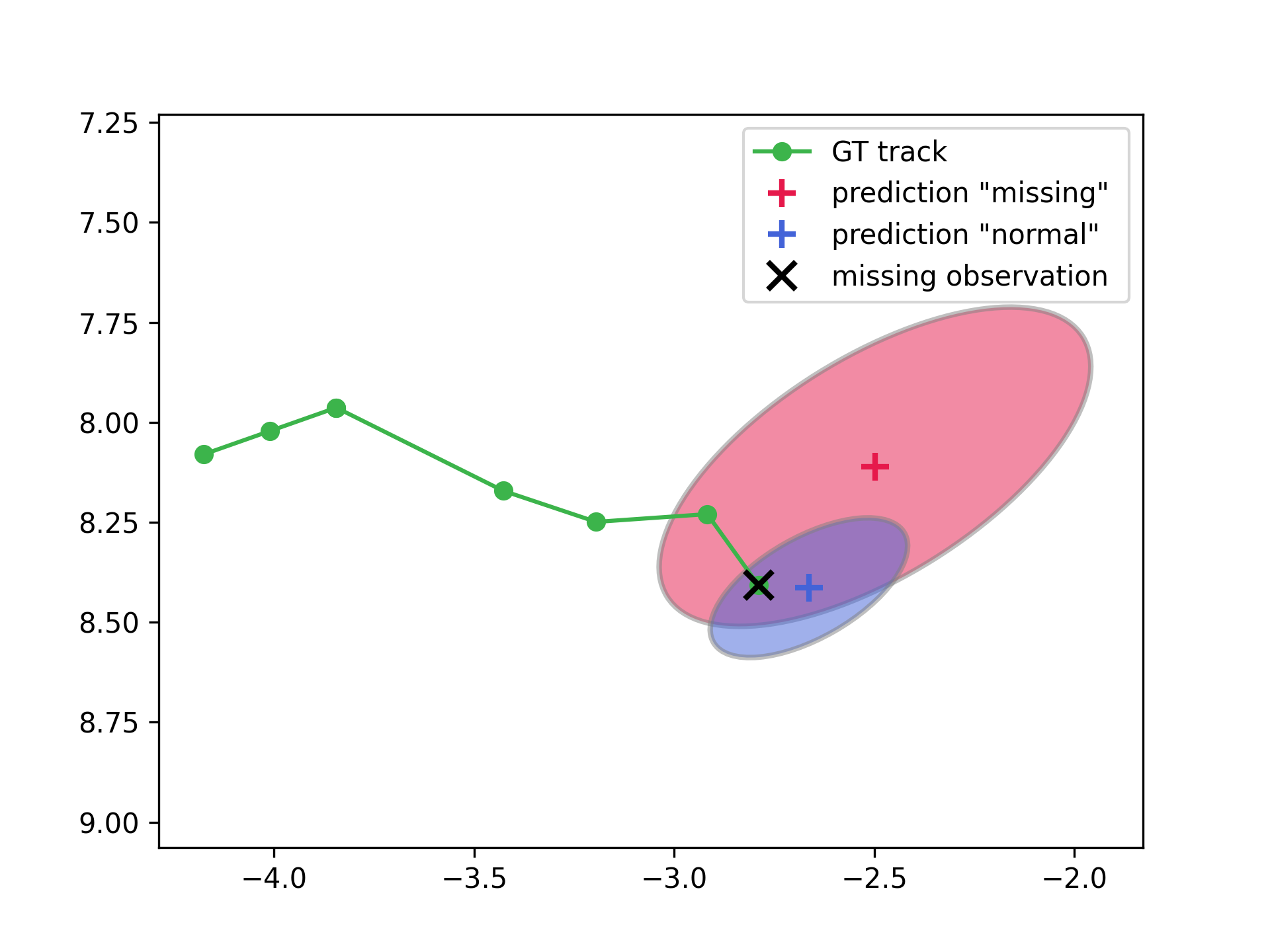}};			
					\node [below of = image1,node distance=2.5cm] (x){$x$/m};				
					\node [left of = image1, node distance=3.3cm,rotate=90] (Y) {$y$/m};
					%\draw[step=0.5cm,gray,very thin] (-4,-4) grid (4.0,4.0);
					\end{scope}									
				\end{tikzpicture} \\				
\end{tabular}
}	
%\vspace{-1.6mm}
\caption{Example predictions for two trajectories from the \emph{ETH} dataset \cite{Pellegrini_ICCV_2009}. The prediction uncertainties with a missing observation are visualized in red. The standard predictions, with providing of the current observation, are shown in blue.}
\label{fig:missing-obervations} 	
\end{figure}
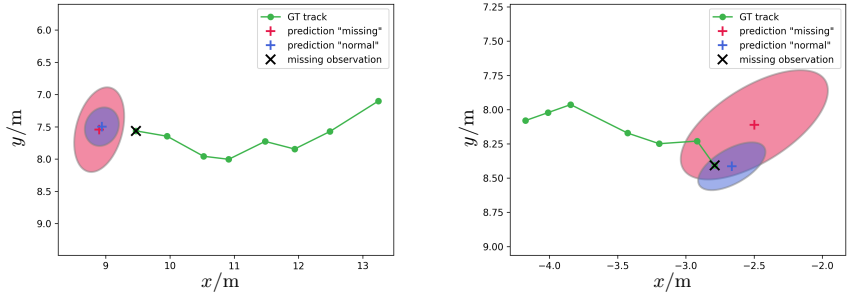\\

\begin{figure}[!ht]
\centering
\resizebox{\textwidth}{!}{
\begin{tabular}{cc}				
				\begin{tikzpicture}
					\begin{scope}[font=\footnotesize]
					\node (image1) at (0,0) {	\includegraphics[height=40mm,width=70mm]{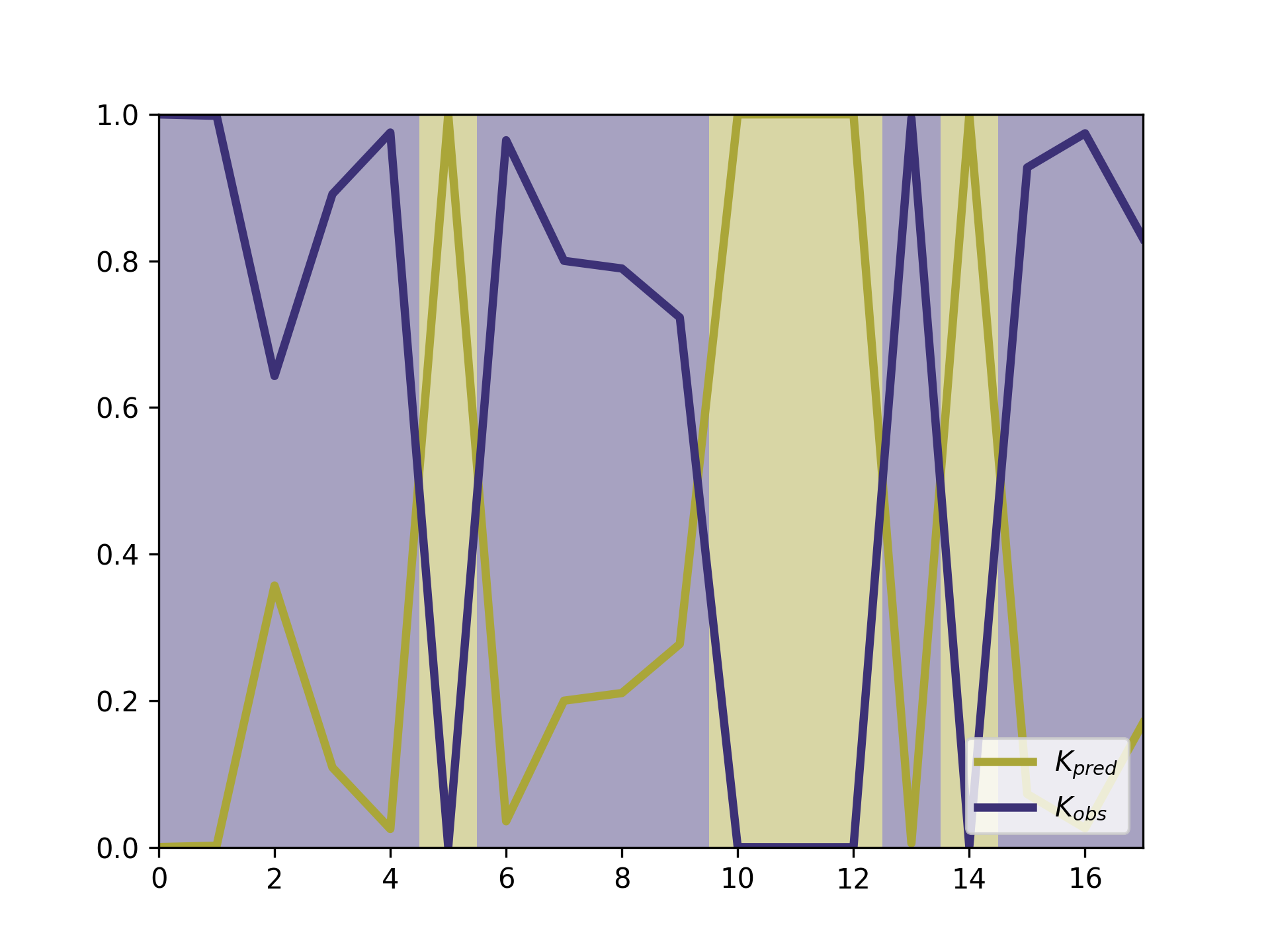}};			
					\node [below of = image1,node distance=2.cm] (x){time steps};				
					\node [left of = image1, node distance=3.2cm,rotate=90] (Y) {pseudo-Kalman gain};
					%\draw[step=0.5cm,gray,very thin] (-4,-4) grid (4.0,4.0);
					\end{scope}									
				\end{tikzpicture} 
			   &		
					\begin{tikzpicture}
					\begin{scope}[font=\footnotesize]
					\node (image1) at (0,0) {	\includegraphics[height=40mm,width=70mm]{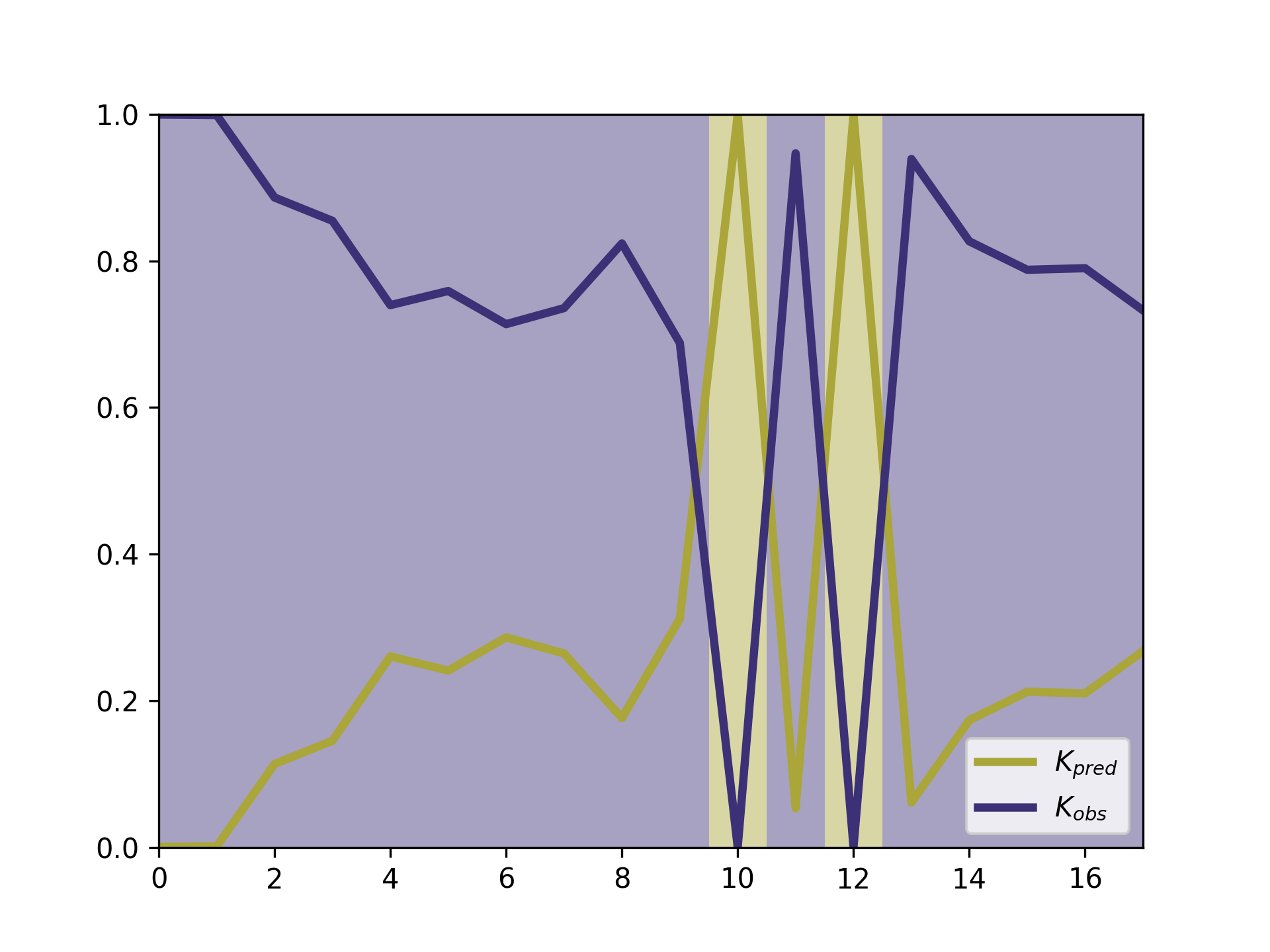}};			
					\node [below of = image1,node distance=2.cm] (x){time steps};				
					\node [left of = image1, node distance=3.2cm,rotate=90] (Y) {pseudo-Kalman gain};
					%\draw[step=0.5cm,gray,very thin] (-4,-4) grid (4.0,4.0);
					\end{scope}									
				\end{tikzpicture}  \\				
\end{tabular}
}	
%\vspace{-1.6mm}
\caption{Visualization of the pseudo-Kalmam gain for low observation noise sequences. The weighting towards trusting the prediction is visualized with dark yellow and towards observations with dark blue. Time steps with missing observations are highlighted with a yellow background.}
\label{fig:weihting-pred-obs} 	
\end{figure}
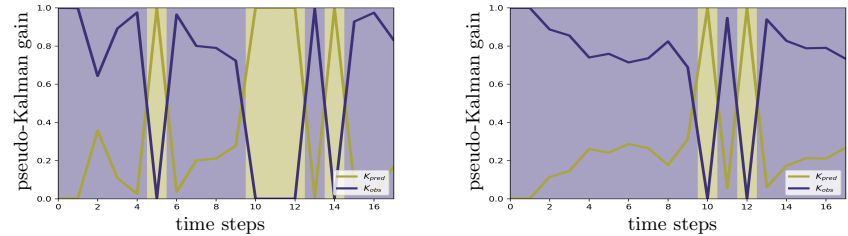

\textbf{Results \& Analysis:} For every experiment, $1000$ noisy trajectories are synthetically generated with a ratio of using $80\%$ for training and $20\%$ for evaluation. The results are summarized in table \ref{result-table-pred-update-cycle}. For comparison, the average displacement error (ADE) is calculated as the average L2 distance between the estimated positions and the ground truth positions. Further, the probability of a missing observation and outlier are varied ($Ber_{}(0.0,1.0)$ and $Ber_{}(0.1,0.9)$). As reference models, a one-to-one RNN-MDN (RNN-($1$to$1$)-MDN), which estimates the true positions stepwise, and an RNN-encoder with an MDN on top (RNN-(encoder)-MDN), which first fully observes the input sequence, are used to generalize from the noisy inputs. The sequence length varies between $8$ and $20$ time steps. In case observations are missing, the reference models receive the predictions from the Prediction-RNN for data imputation.
 Compared to linear interpolation, a better performance is achieved in the experiments and better comparability to the Prediction-Update cycle is guaranteed. These results show that the proposed Prediction-Update-RNN can better handle outliers and missing observations. Even in the experiments without outliers, the achieved result is better. Due to provided binary-coded masking patterns, the approach learns to ignore the placeholder inputs and to fully trust the predictions. Missing observations lead to increasing prediction uncertainties. Thus the model corrects the position estimates by relying more strongly on the new observations. These effects are visualized in figure \ref{fig:missing-obervations} and \ref{fig:weihting-pred-obs}. Figure \ref{fig:weihting-pred-obs} shows the pseudo-Kalman gain for a low observation noise sequence. The weighting towards trusting the prediction is visualized with dark yellow ($K_{pred}=(1-K_{obs})$) and correspondingly $K_{obs}$ with dark blue. Following this color scheme, time steps with missing observations are highlighted with a dark yellow background. It is clearly visible how the approach relies only on the predicted position to estimate the posterior position when the observation is replaced with a placeholder value. The ability of the Prediction-RNN to capture the increased prediction uncertainty when observations are missing is depicted in figure \ref{fig:missing-obervations} for an example trajectory of the \emph{ETH} dataset \cite{Pellegrini_ICCV_2009}. The ground truth trajectory for conditioning is shown in green. Missing observations are marked with a cross. The covariance ellipses capture the $3\sigma$ area around the predicted position. The predicted position varies reasonably around the prediction from the fully observed trajectory. Besides demonstrating the Prediction-RNN ability, this example shows the noise present in the ground truth data. Since RNNs can generalize to produce smooth trajectories, neglecting such noise levels in order to focus on prediction comparison, seems, on the one hand, to be reasonable. On the other hand, evaluating the filtering performance by not considering the noise or discretization artifacts in the underlying trajectories data is not adequate. Further, the position estimation error naturally influences prediction performance because the position estimate often serves as the reference point for long-term prediction.  

%In summary, the presented Prediction-Update-RNN can better handle missing observations and outliers present in time-series data. Like in the prediction and update step of a Kalman filter, the RNNs exchange their prior and posterior estimate, but the estimates depend on the whole history of inputs. This study on synthetically generated data shows that by exploiting the connections between different views on inference problems, perspectives on overcoming respective limitations can be gained.

\section{Conclusion}
\label{sec:conclusion}
In this paper, an RNN-based Prediction-Update cycle has been presented. The model enables improved handling of missing observations and outliers present in time-series data. The model abilities were shown on synthetic data reflecting prototypical pedestrian maneuvers. By iteratively exchanging the estimates of two separated RNNs and providing a binary-coded missing pattern, the model can learn to trust the prior estimates or rely more strongly on the current observations.    

\bibliographystyle{splncs04}
\bibliography{Becker_CAIP_2021}

\end{document}